\documentclass{article}

\usepackage[final]{nips_2018}

\usepackage[utf8]{inputenc} 
\usepackage[T1]{fontenc}    
\usepackage{hyperref}       
\usepackage{url}            
\usepackage{booktabs}       
\usepackage{amsfonts}       
\usepackage{nicefrac}       
\usepackage{microtype}      
\usepackage{amsmath}
\usepackage{xcolor}

\usepackage{subfig}
\usepackage{graphicx}
\usepackage{multirow}
\usepackage{amssymb}
\usepackage{ulem}
\usepackage{placeins}

\newcommand\todo[1]{}

\usepackage{array,multirow}
\usepackage{listings}
\usepackage{booktabs}

\title{Compression and Localization in Reinforcement Learning for ATARI Games}

\author{
  Joel Ruben Antony Moniz \thanks{Indicates equal contribution.} \\
  Carnegie Mellon University\\
  \texttt{jrmoniz@andrew.cmu.edu} \\
  \And
  Barun  Patra ${^{*}}$ \\
  Carnegie Mellon University\\
  \texttt{bpatra@andrew.cmu.edu} \\
  \And
  Sarthak Garg ${^{ *}}$ \\
  Carnegie Mellon University\\
  \texttt{sarthakg@andrew.cmu.edu} \\
}

\begin{document}

\maketitle

\begin{abstract}

Deep neural networks have become commonplace in the domain of reinforcement learning, but are often expensive in terms of the number of parameters needed. While compressing deep neural networks has of late assumed great importance to overcome this drawback, little work has been done to address this problem in the context of reinforcement learning agents. This work aims at making first steps towards model compression in an RL agent. In particular, we compress networks to drastically reduce the number of parameters in them (to sizes less than 3\% of their original size), further facilitated by applying a global max pool after the final convolution layer, and propose using Actor-Mimic in the context of compression. Finally, we show that this global max-pool allows for weakly supervised object localization, improving the ability to identify the agent's points of focus.

\end{abstract}

\section{Introduction}

Deep Reinforcement Learning (Deep RL) has lately seen increasing applications to real world problems: in robotics (such as by \citet{levine2016end}), real-time bidding (such as by \citet{jin2018real}) and dialog generation (such as by \citet{li2017adversarial}). With increasing real-world applicability, making light-weight Deep RL agents is increasingly important. While compressing networks has widely been adopted in domains where Deep Learning techniques are often applied, such as computer vision (for example, in \citet{han2015deep}), exploring the compression of networks in Deep RL is relatively less explored.

In this work, we aim to perform model compression in the context of a Deep RL agent. We propose using a formulation similar to Actor-Mimic \citep{parisotto2015actor} to distill the knowledge of a trained expert into a student network, which has significantly fewer parameters and a lower computational cost. We validate this using a test bed commonly used in RL: ATARI games using the ALE environment \citep{bellemare2013arcade}, and use a Deep Q-Network \citep{DQN} as our expert model. This is to the best of our knowledge one of the first times model compression has been performed in the setting of a Deep RL agent.

While one of the ways in which we reduce parameters in our student network involves directly reducing the number of feature maps, we also explore improving the parameter efficiency using a global max pool after the final convolutional layer. In addition to drastically reducing the required number of parameters (now utilizing fewer than 3\% of the parameters required by the original expert network), this model formulation serves as an implicit form of attention.
The global max pool further allows us to visualize object localizations predicted by the network, since it encourages a form of weakly supervised object detection \citep{oquab2015object}. In particular, we find that the global max-pooling often induces the localization of the important entities in the game, and serves as a window to interpret what the agent is focusing on while making an action. Further, this localization allows us to potentially glean what strategies the agent seeks to take.

\section{Proposed Method}

In order to significantly cut down on the parameters, we simplify the architecture by halving the number of feature maps across all convolutional layers in our Deep Q-Network, and then use a global max-pool between the last convolutional layer and the first fully connected layer. These light-weight  networks are significantly harder to directly optimize using the traditional Q-Learning objective. We use knowledge distillation as a form of imitation learning to train these parameter-efficient networks. This section describes the details of the aforementioned methods.

\subsection{Pooling as an attention proxy}
\cite{xu2015show} demonstrated the effectiveness of using visual attention. The basic idea behind visual attention is to learn a probability distribution over the features, which can be interpreted as a measure of the relevance of the feature for prediction. The features are weighted by this relevance before being passed into downstream prediction networks. Visual attention has been shown to work well in 3D environments \citep{chaplot2017gated} and in Atari game environments \citep{sorokin2015deep}. 

The original \citet{DQN} formulation has around 4 million parameters, a majority of which come from the fully connected layers after the final convolutional feature maps. We believe that these layers are overparameterized. This problem can be elevated by using attention modules to select the most important features and passing only those to the fully connected layers. For example adding a simple dot product-based attention over feature maps would allow one to cut down the parameters to only around $392k$ parameters.
Furthermore the resulting attention maps enable us to localize the important features allowing for better feature visualization.

However, these attention vectors themselves often add their own set of parameters, and the accompanying effort required to optimize these parameters. To avoid this, we propose simply applying a global max-pool at the end of the convolutional pipeline. In addition to reducing the number of parameters, applying a global max-pool encourages localization purely by weak supervision. 

An important difference of our proposed formulation is that we no longer have an explicit attention probability mask, relying solely on a global pool in its place; in addition, we max over the features being attended to, as opposed to a sum weighted by the predicted attention probability mask \citep{xu2015show}. We do this inspired by work in the domain of Weakly Supervised learning, such as \cite{oquab2015object}, in which the authors show that using a max operation when classifying images acts as a form of weak supervision for the localization of the classified objects, and networks trained in this manner successfully learn to localize important objects in spite of not having ground-truth labels at train time.

Since the global max-pool operation allows only the features with the highest activation across the whole spatial dimension in a given feature map, the network is forced to assign the highest activations to the most relevant entities. The global max pooling thus induces the localization of the entities important to the game. As a result, visualizing these pre-max-pool activation maps allows us to see what the network considers the most important entities. 
Since the agent also has to strategize about how it aims to play the game, these visualizations potentially reveal elements of the adopted strategy.

\subsection{Parameter reduction using Actor-Mimic}

\cite{hinton2015distilling} showed that it might be easier for a small network to learn to mimic the predictions of a large network or an ensemble of networks trained on a task as compared to learning that task from scratch. 

An important objective of this work is to investigate the ability to distill the knowledge of an RL agent into a highly compressed student network. This is akin to imitation learning, in which, given an expert, a student network is trained to mimic the expert's policy. We use the uncompressed, trained deep Q-Network as our experts, and train light-weight, compressed networks as the student networks.

\cite{parisotto2015actor} effectively used the above idea in the domain of deep reinforcement learning, to train an actor agent to play multiple games by mimicking the respective experts. The student DQN can mimic the expert by naively regressing it's output against the expert's Q values. However the range of Q values varies widely with the input state, leading to training instability. 
\citet{parisotto2015actor} circumvent this problem by converting the Q values to a probability distribution over the actions, and then matching the student's and the expert's policy distributions. 

We employ the same technique as above in our method. Specifically, the Q values are first converted into a probability distribution as follows
\begin{equation*}
\pi(a | s) = \frac{e^{\tau^{-1}Q(s, a)}}{\sum_{a^{'} \in \mathcal{A}}e^{\tau^{-1}Q(s, a^{'})}}
\end{equation*}
where $\tau$ is the temperature parameter. For a state $s$, given the expert policy $\pi_{E}(a | s)$, the student network (S) is trained to minimize the KL divergence between $\pi_{E}(a | s)$ and $\pi_{S}(a | s)$. The state action $(s, a)$ pairs used for training are sampled from the student network's learned policy. 
\section{Experimental Setup}
\begin{table}[!t] 
\centering 
\caption{Comparing the performance of the student network with the expert network}
\vspace{-2mm}
\begin{tabular}{@{}lccccccc@{}} \\
\toprule
\textbf{Model} & \textbf{SpaceInvaders} & \textbf{BeamRider} & \textbf{Breakout} & \textbf{Boxing} & \textbf{Enduro} & \textbf{Pong} & \textbf{Seaquest} \\
\midrule
\multirow{2}{*}{Expert} & 1388.2 & 8520.1  & 276.5  & 95.76  & 1428.6 & 20.9  & 5769.4  \\
& $\pm$ 351.7 & $\pm$ 2968.6 & $\pm$ 140.9 & $\pm$ 3.64 & $\pm$ 325.9 & $\pm$ 0.25 & $\pm$ 1760.1 \\
\midrule
\multirow{2}{*}{Student} & 1325.4 & 5602 & 375.7 & 96.5 & 1402.5 & 20.8 & 4860.9 \\
& $\pm$ 504.6 & $\pm$ 3336.1 & $\pm$ 67.4 & $\pm$ 4.8 & $\pm$ 342.8 & $\pm$ 0.49 & $\pm$ 1975.3 \\
\bottomrule
\end{tabular}
\label{tab:main-results}
\end{table}

\textbf{Games:} We train and compare the performance of the parameter heavy experts and lightweight student networks on 6 different games: Space Invaders, Breakout, Seaquest, Beamrider, Pong, Boxing and Enduro.

\textbf{Expert networks} are trained using the vanilla DQN objective and use the architecture proposed by \citet{DQN}. The training hyperparameters are roughly similar to \citet{DQN}\todo{, for details refer to Add link to appendix}.

\textbf{Student networks} are trained by minimizing the KL divergence between the expert policy and the student policy. The features of the last convolutional layer in the expert networks are of dimension 7x7x64, 64 maps each of size 7x7. The student networks reduce the dimensionality of these features by halving the number of feature maps from 64 to 32 and doing a spatial max over 7x7 features to 1 feature. Thus, the final convolutional layer in the student networks is of dimension 1x1x32.

\textbf{Localization} maps are generated by using the 7x7 activation maps of the last convolutional layer of the student networks before spatial max pooling. The values in these activation maps are normalized to the range $(0, 1)$ and the maps are then upsampled to the original game frame dimensions using bilinear interpolation. The normalized and rescaled activation maps are superimposed over frames to generate visualizations localizing the important entities and strategies of the games.

\textbf{Training Methodology:} The expert model is trained for 10 million iterations and is evaluated on 10 episodes after every 0.5 million iterations. The best expert model is selected based on this evaluation and is used for training the student network. The best expert model is also run again for 100 episodes and the mean and std deviation of rewards are recorded as the final performance of the expert. 
The student networks are trained for 20 million iterations and is evaluated on 10 episodes after every 0.5 million iterations. The best student model based on this evaluation is run again for 100 episodes and the mean and std deviation of the rewards are recorded as the final performance of the student.

\section{Results and Analysis}
\begin{table}[!t] 
\centering 
\caption{A comparison of the number of parameters used in various settings}
\label{tab:params}
\vspace{-2mm}
\begin{tabular}{@{}cccc@{}} \\ 
\toprule
\textbf{Expert} & \textbf{Ours} & \textbf{Max Same} & \textbf{None Halved} \\
\midrule
1.68M & 43.1K & 115.88K & 829.51K \\
\bottomrule
\end{tabular}
\end{table}
\subsection{Results}
We summarize in Table \ref{tab:main-results} the performance of the expert networks and compare it with that of the networks with half the number of feature maps and global max pooling trained with knowledge distillation using Actor-Mimic (i.e., the student networks). Despite the student networks containing about only 3\% of the parameters of the expert networks, they show performance reasonably close to the expert networks on a wide range of games (Space Invaders, Breakout, Boxing, Enduro and Pong). We primarily attribute this to knowledge distillation which makes it significantly easier for smaller networks to learn as opposed to learning from scratch. Specifically, the expert networks take 1.69M parameters, while the student network requires just 43.1k.

\subsection{Feature Map Visualization}
\begin{figure*}[!thb]
\centering
\begin{minipage}{0.3\textwidth}
 \centering
 \includegraphics[width=1.\textwidth]{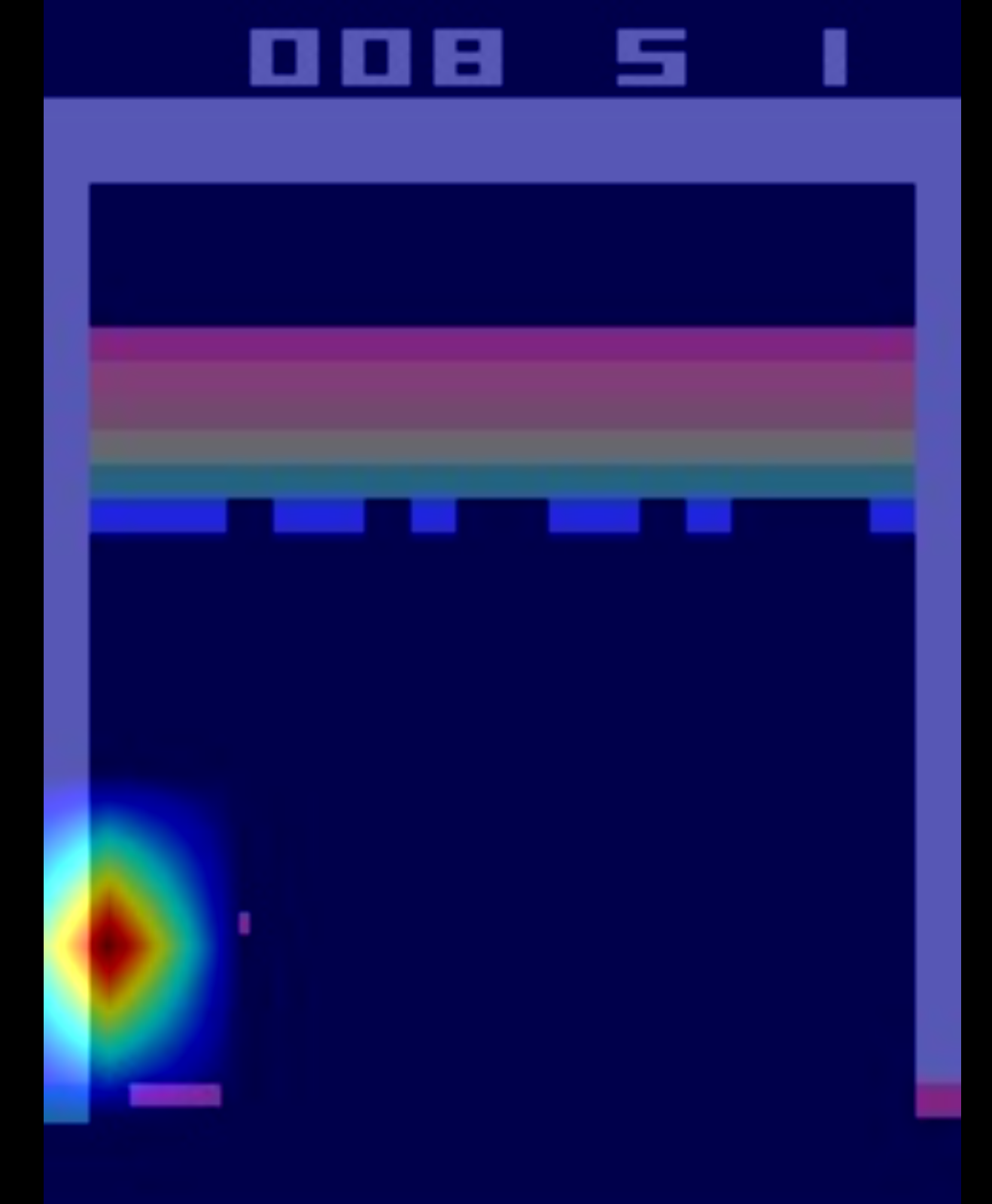}
 \caption*{Breakout}
\end{minipage}
\hfill
\begin{minipage}{0.3\textwidth}
 \centering
 \includegraphics[width=1.\textwidth]{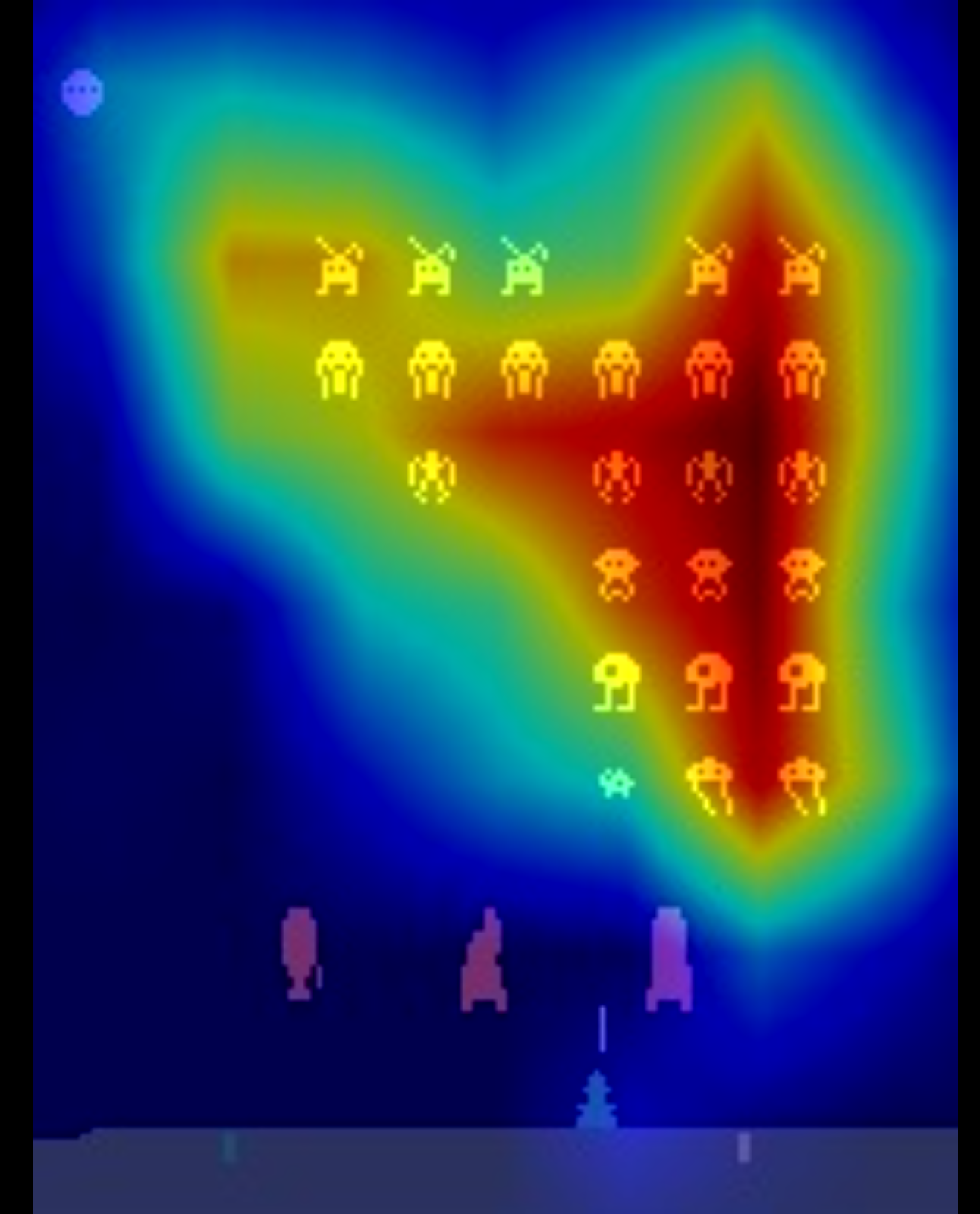}
 \caption*{Space Invaders}
\end{minipage}
\hfill
\begin{minipage}{0.3\textwidth}
 \centering
 \includegraphics[width=1.\textwidth]{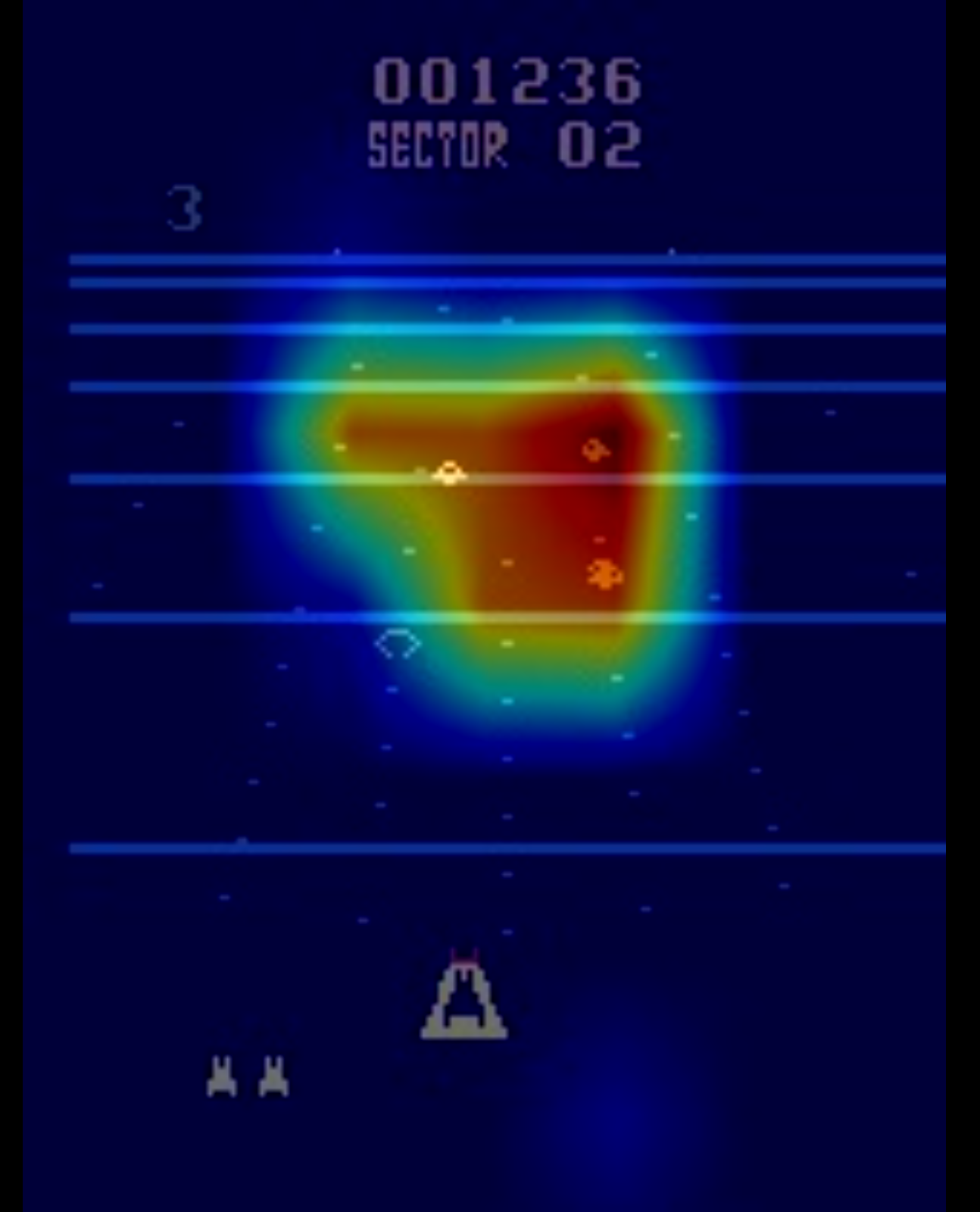}
 \caption*{Beam Rider}
\end{minipage}

\hspace{-1em}
\caption{Strategies adopted by the model}
\label{fig:strategy}

\begin{minipage}{0.3\textwidth}
 \centering
 \includegraphics[width=1.\textwidth]{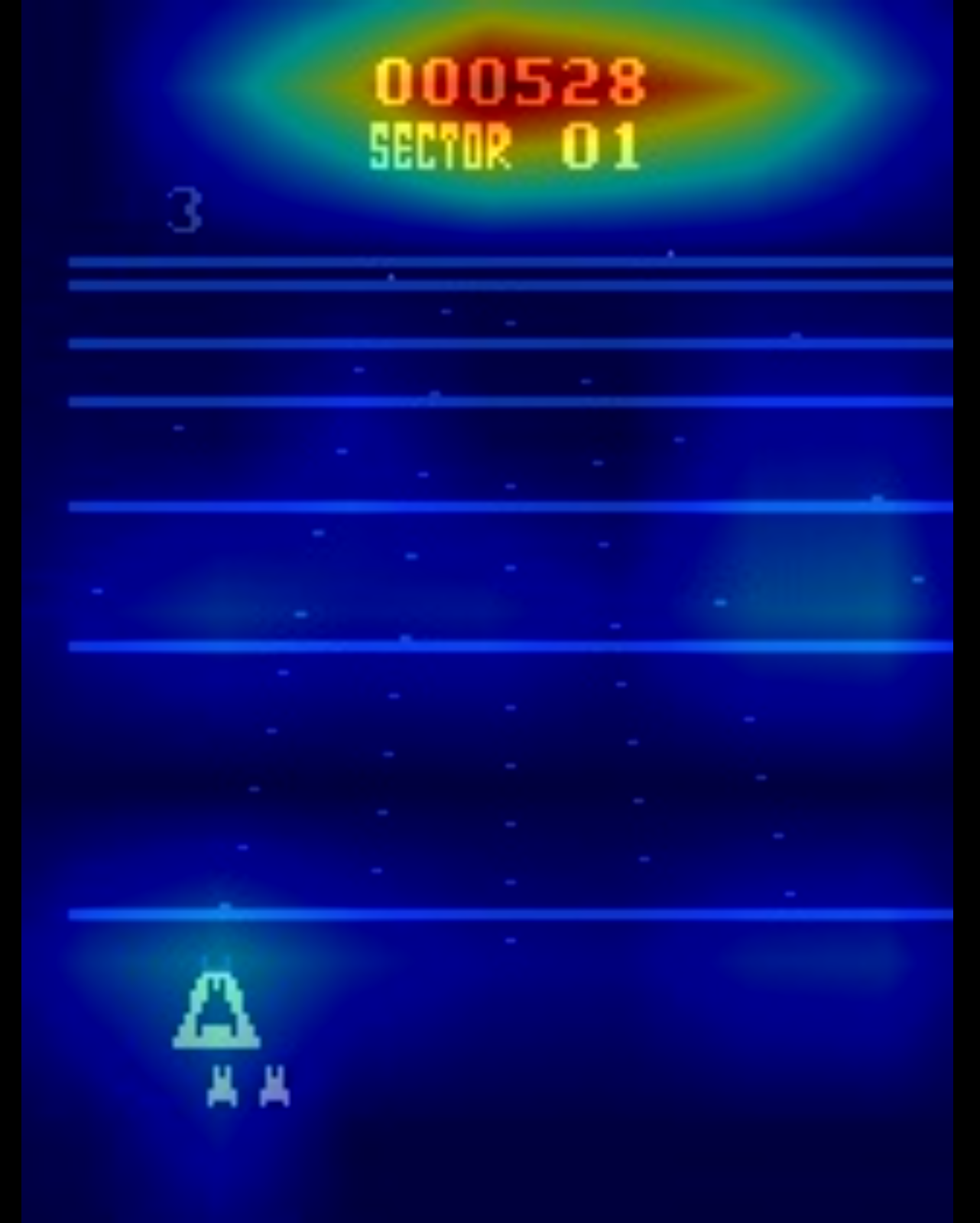}
 \caption*{Beam Rider}
\end{minipage}
\hfill
\begin{minipage}{0.3\textwidth}
 \centering
 \includegraphics[width=1.\textwidth]{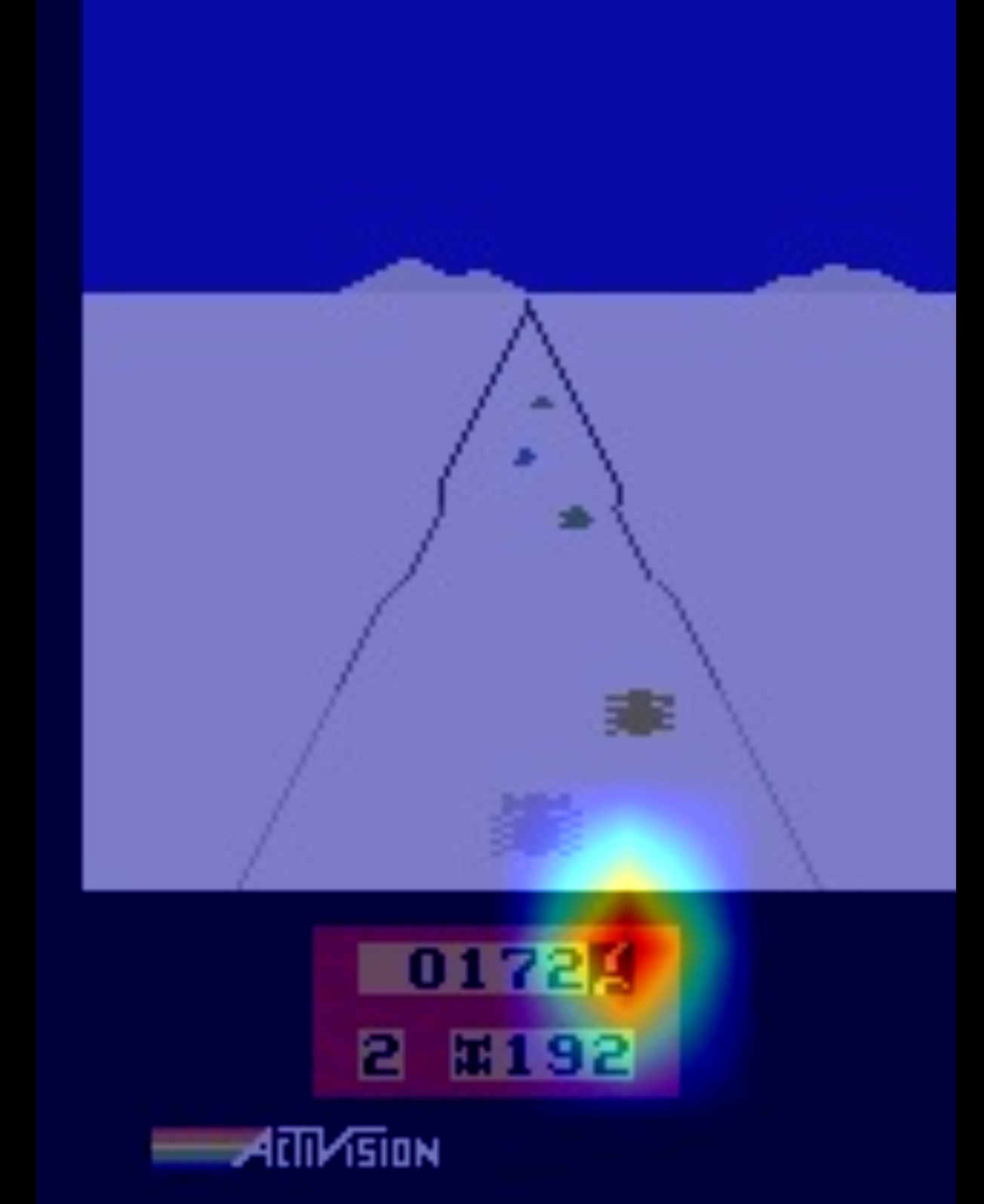}
 \caption*{Enduro}
\end{minipage}
\hfill
\begin{minipage}{0.3\textwidth}
 \centering
 \includegraphics[width=1.\textwidth]{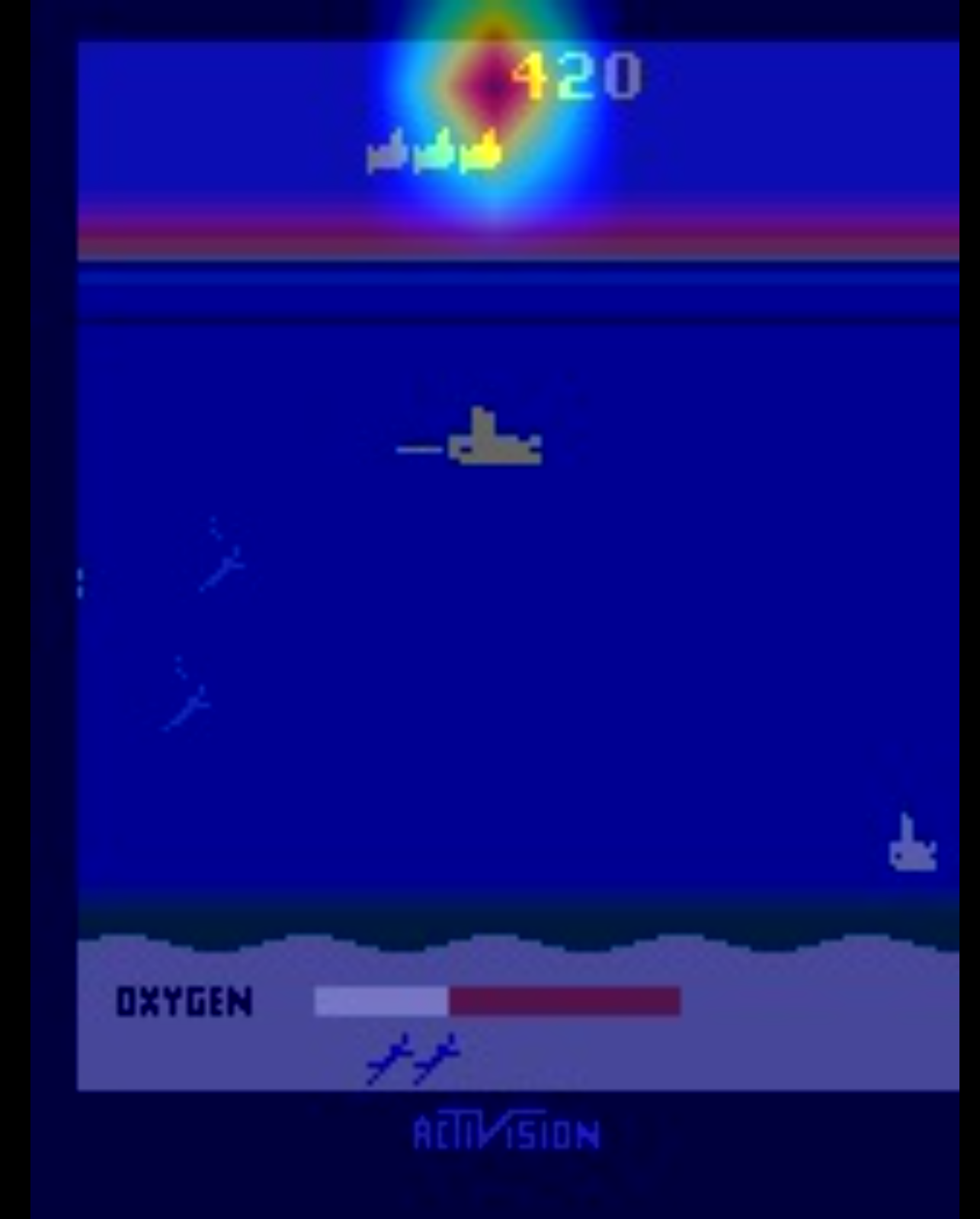}
 \caption*{Seaquest}
\end{minipage}

\hspace{-1em}
\caption{Feature maps confused by the score}
\label{fig:confusion}
\end{figure*}

\begin{figure*}[tb]
\centering
\begin{minipage}{0.31\textwidth}
 \centering
 \includegraphics[width=1.\textwidth]{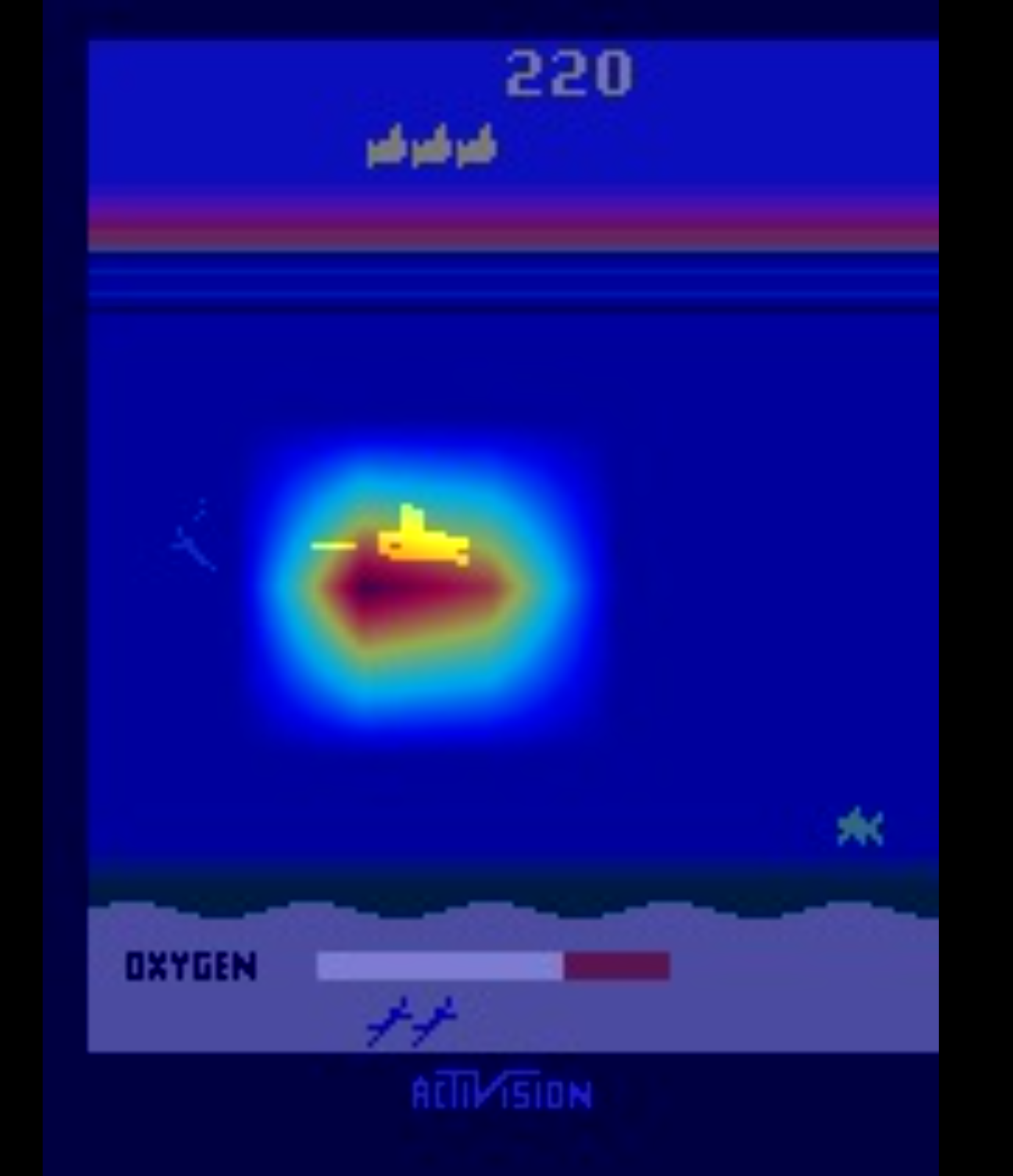}
 \caption*{Focus on the submarine}
\end{minipage}
\hfill
\begin{minipage}{0.31\textwidth}
 \centering
 \includegraphics[width=1.\textwidth]{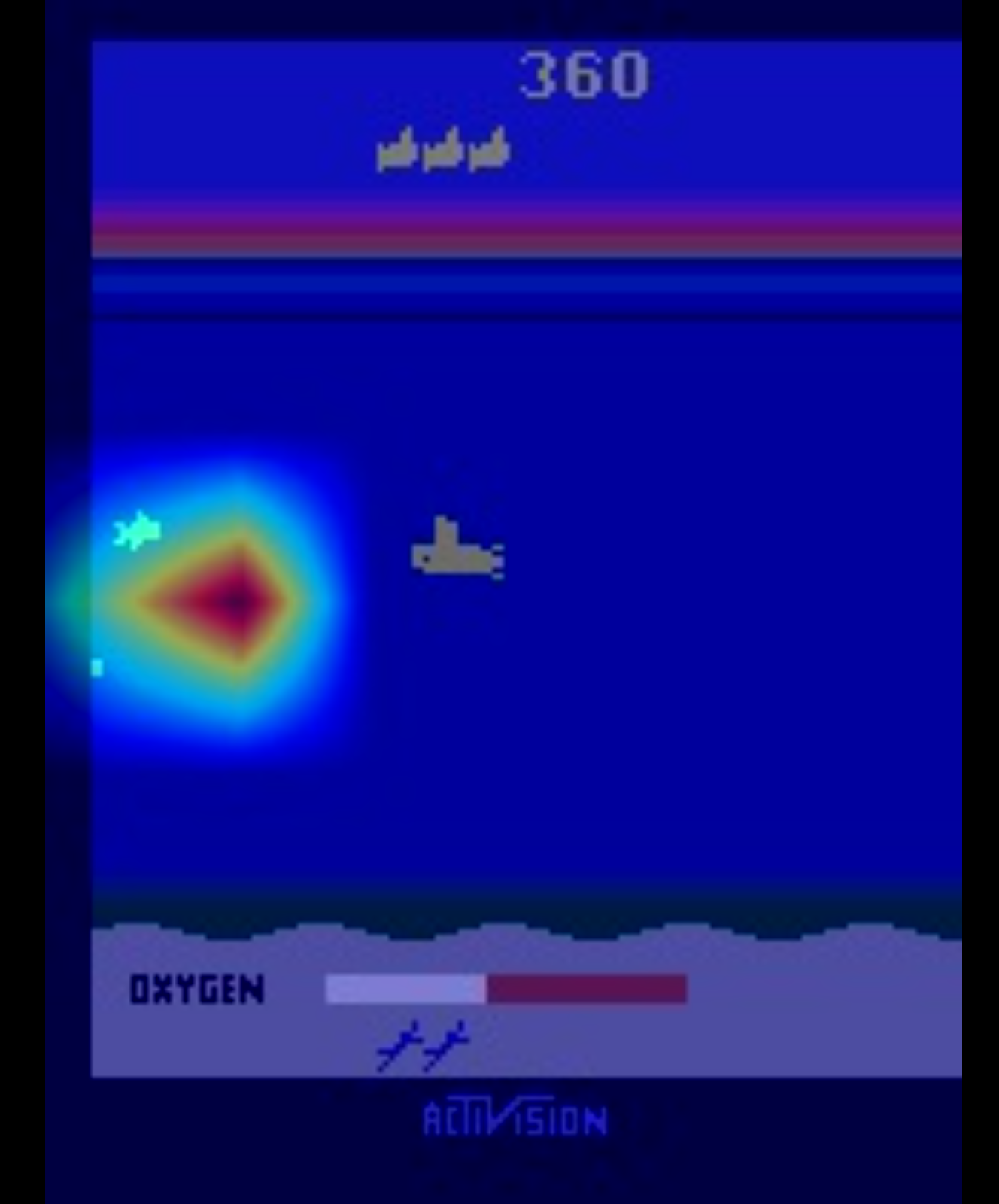}
 \caption*{Focus on fish}
\end{minipage}
\hfill
\begin{minipage}{0.31\textwidth}
 \centering
 \includegraphics[width=1.\textwidth]{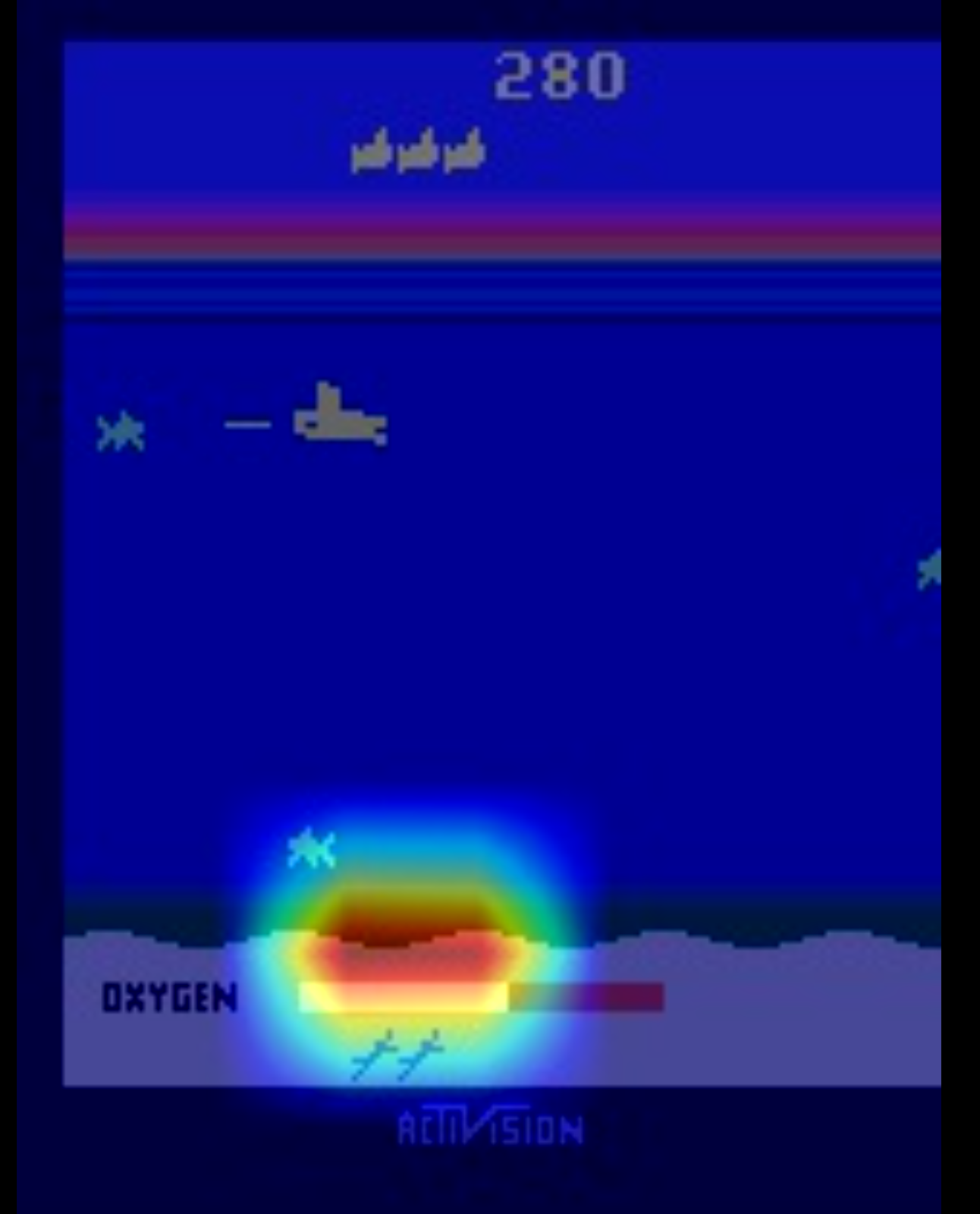}
 \caption*{Focus on Oxygen}
\end{minipage}
\hspace{-1em}
\caption{Maxpool maps for Seaquest}
\label{fig:seaquest}

\begin{minipage}{0.31\textwidth}
 \centering
 \includegraphics[width=1.\textwidth]{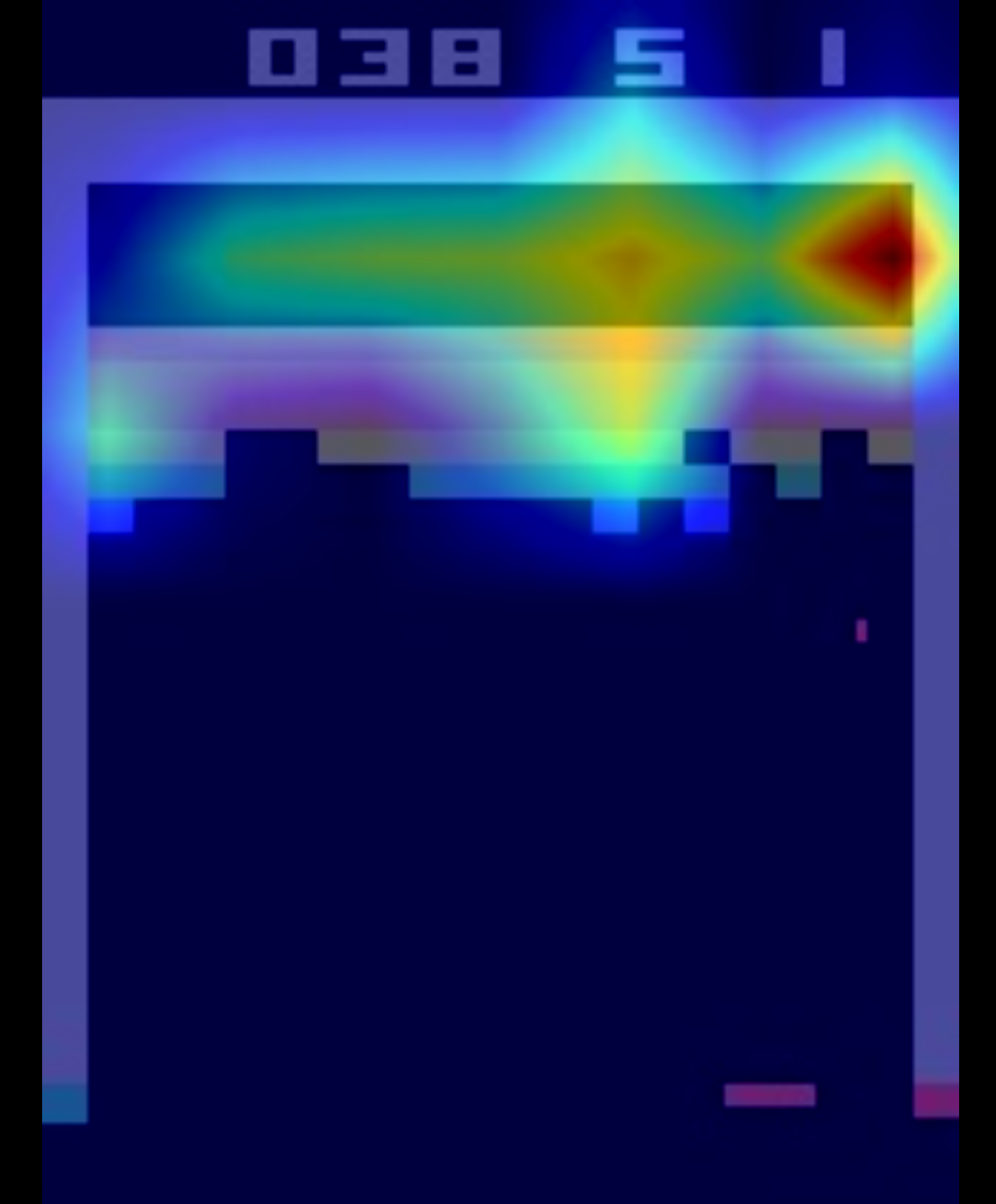}
 \caption*{Focus on bricks}
\end{minipage}
\hfill
\begin{minipage}{0.31\textwidth}
 \centering
 \includegraphics[width=1.\textwidth]{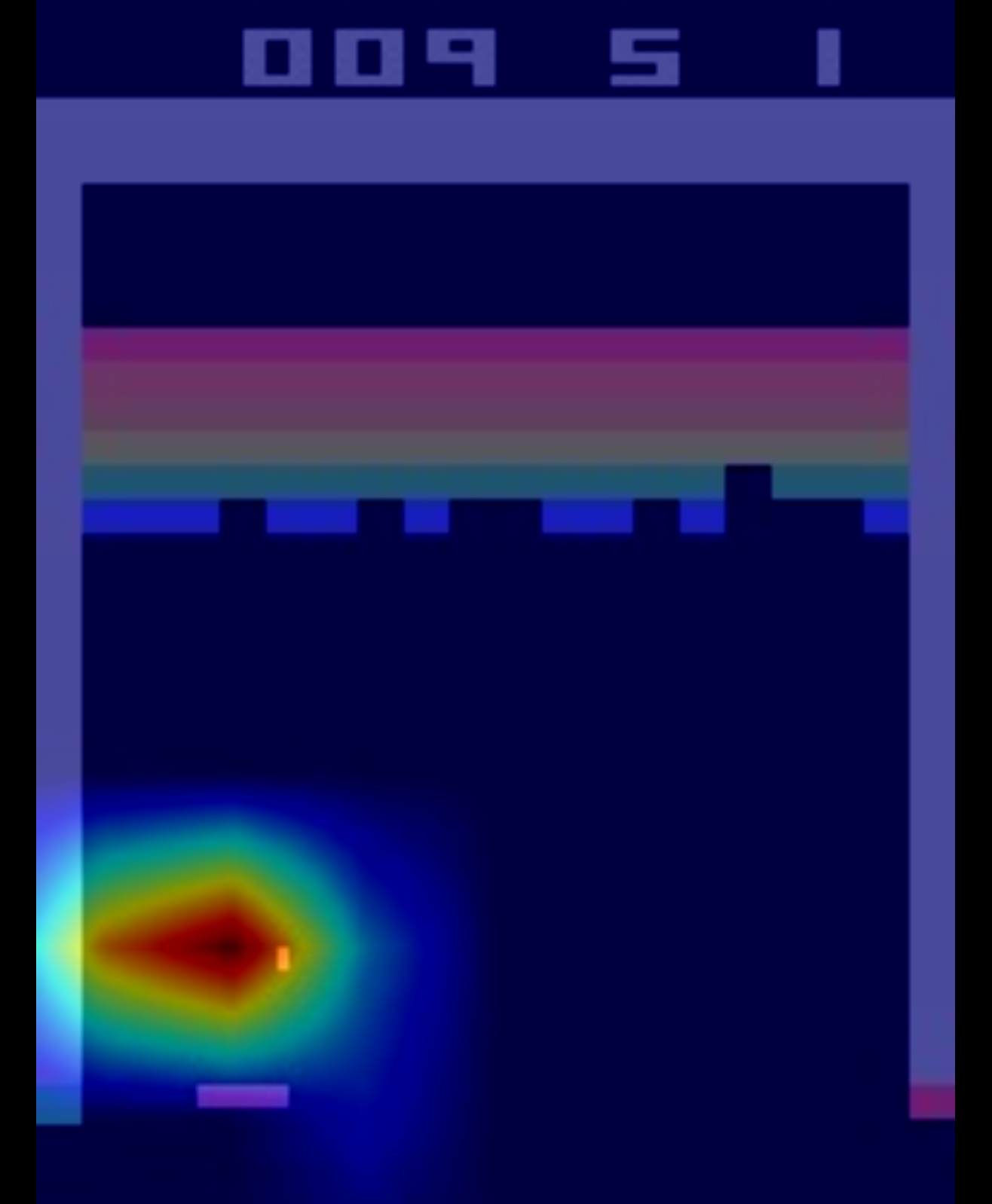}
 \caption*{Focus on ball}
\end{minipage}
\hfill
\begin{minipage}{0.31\textwidth}
 \centering
 \includegraphics[width=1.\textwidth]{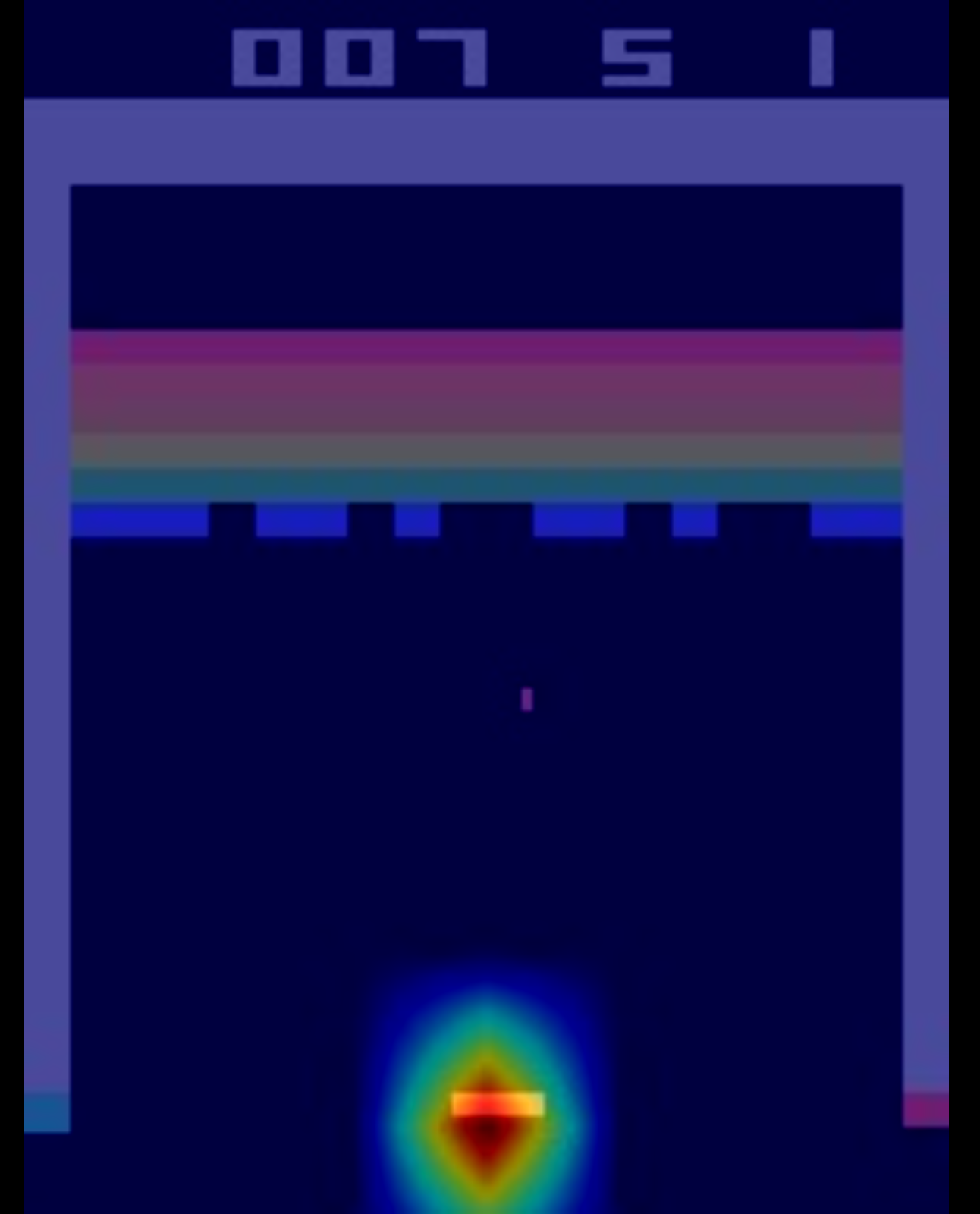}
 \caption*{Focus on paddle}
\end{minipage}
\hspace{-1em}
\caption{Maxpool maps for Breakout}
\label{fig:breakout}

\begin{minipage}{0.31\textwidth}
 \centering
 \includegraphics[width=1.\textwidth]{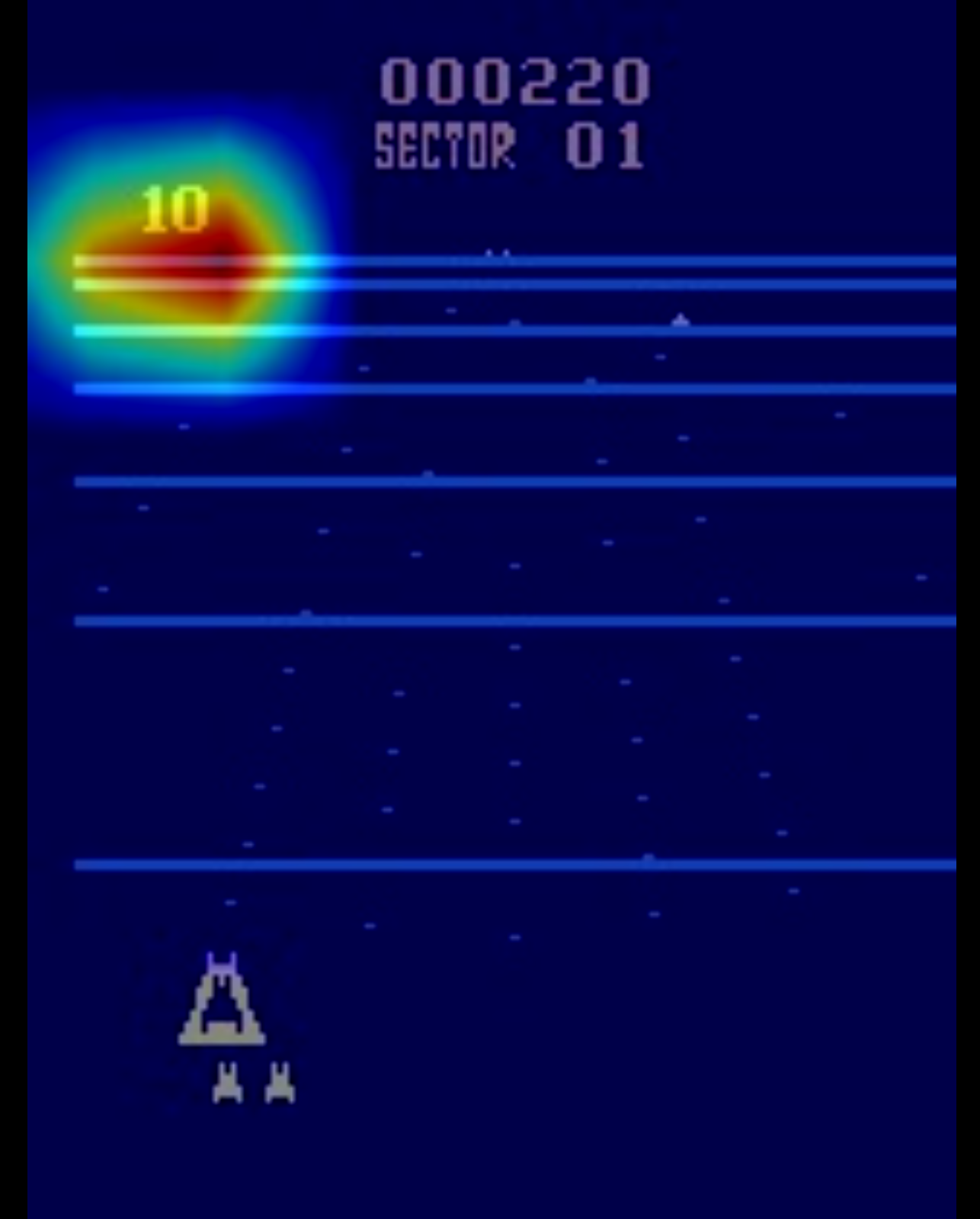}
 \caption*{Focus on distance}
\end{minipage}
\hfill
\begin{minipage}{0.31\textwidth}
 \centering
 \includegraphics[width=1.\textwidth]{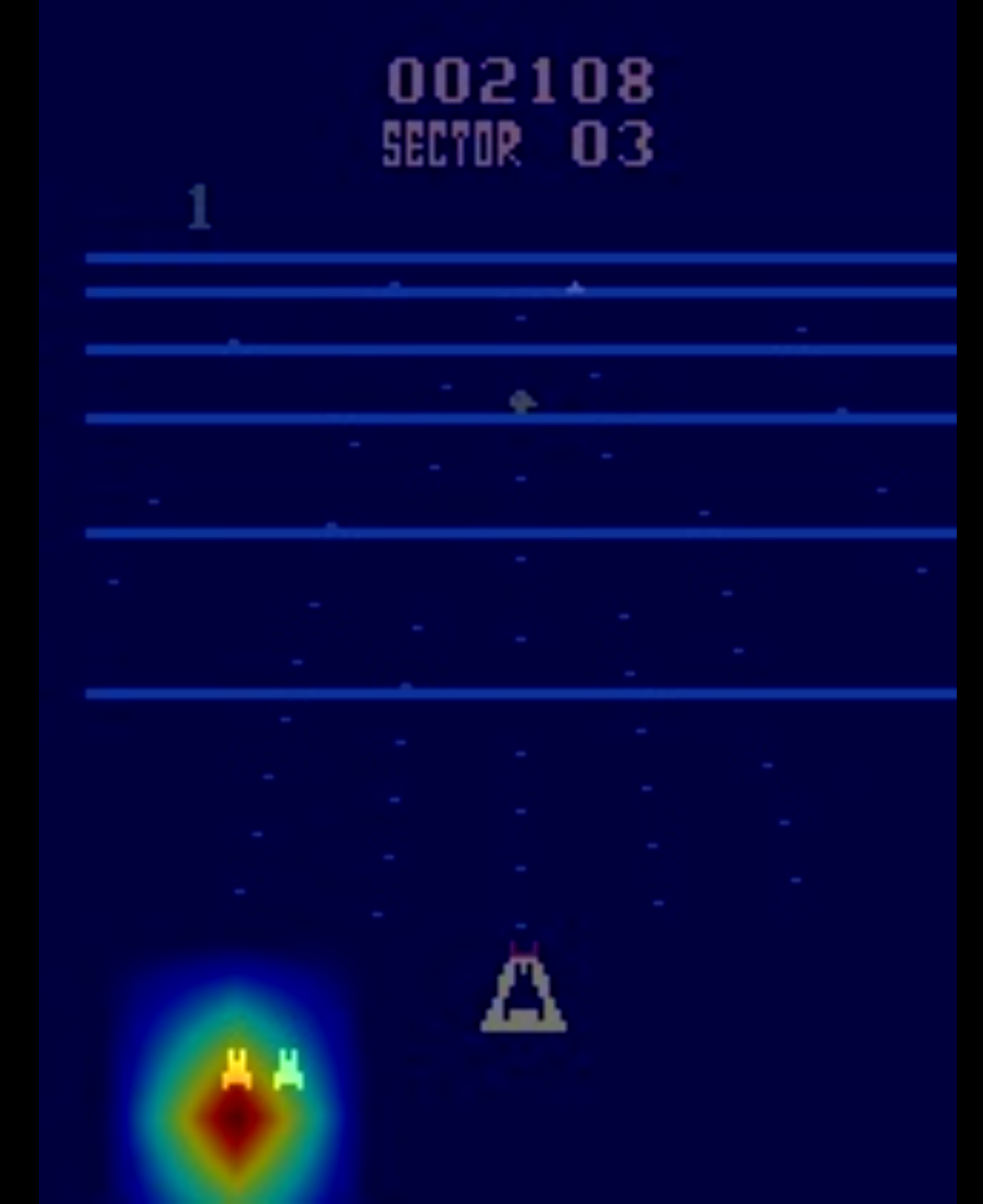}
 \caption*{Focus on lives}
\end{minipage}
\hfill
\begin{minipage}{0.31\textwidth}
 \centering
 \includegraphics[width=1.\textwidth]{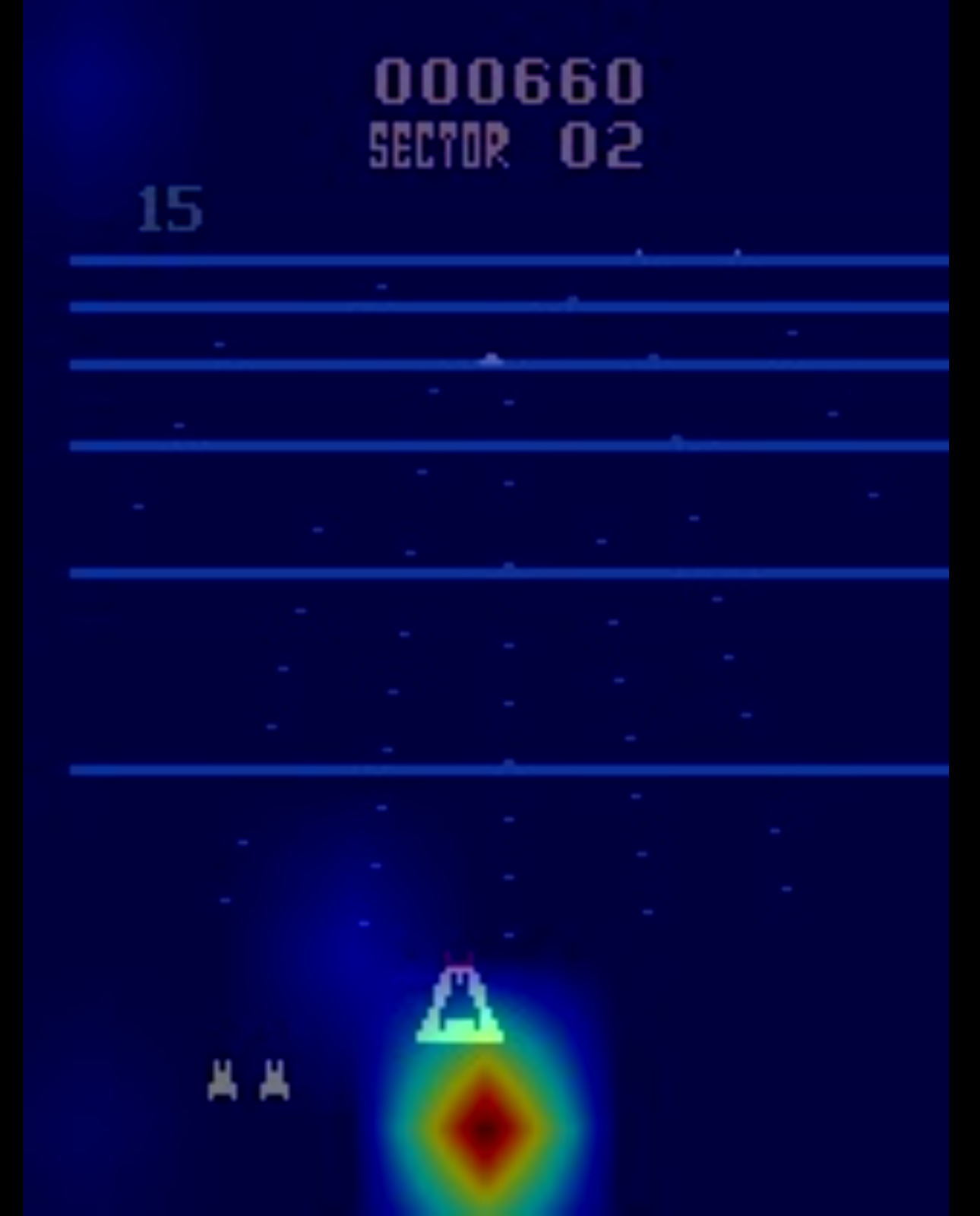}
 \caption*{Focus on agent}
\end{minipage}
\hspace{-1em}
\caption{Maxpool maps for BeamRider}
\label{fig:beamrider}

\end{figure*}

Figures \ref{fig:seaquest}, \ref{fig:breakout} and \ref{fig:beamrider} show the weak localization maps obtained before the global max pooling layer. In the case of Seaquest (Figure \ref{fig:seaquest}), we find that the model focuses on the submarine, fish, and oxygen tank. For Breakout (Figure \ref{fig:breakout}), the model focuses on the remaining bricks, the ball and the paddle. Likewise, in the case of BeamRider (Figure \ref{fig:beamrider}), the model focuses on the distance, the number of lives left, and tracking the current agent, along with the enemies (Figure \ref{fig:strategy}, middle image). Thus, the feature maps focus on the important aspects necessary for good performance in the respective games. This shows that a max pool is generally sufficient to extract relevant features, and can be used in place of a traditional attention module, with the gains of parameter reduction.

Another benefit obtained from visualizing these feature maps allows us to understand the strategies learned by the model. For example, consider Figure \ref{fig:strategy}. The first figure shows a commonly used strategy by the model: trying to hit the lower bottom corner so that the ball can tunnel through to obtain a high score. Similarly, the second and third figures highlight the presence of enemies, with a significant weight given to regions of high densities. We observe that the model does focus on them, trying to eliminate the dense regions, since they pose a greater threat to the model.

However, these maps can sometimes confuse correlation with causation. One of the most commonly observed trends is that these maps end up focusing strongly on the score. This is because the score changes with events of the game, even though the event of the score changing in itself has little impact on the quality of the game played.

\subsection{Ablation Study}
\begin{table}[!t] 
\centering 
\caption{Ablation study analyzing the effect of Actor-Mimic, feature halving and global maxpool.}
\vspace{-2mm}
\begin{tabular}{@{}lcc@{}|@{}cc@{}|@{}cc@{}|@{}cl@{}} \\ 
\toprule
\multirow{3}{*}{\textbf{Games}} & \multicolumn{4}{c}{\textbf{Knowledge Distillation}} & \multicolumn{4}{c}{\textbf{No KD}} \\
\cline{2-9}
& \multicolumn{2}{c}{\textbf{Max}} & \multicolumn{2}{c}{\textbf{None}} & \multicolumn{2}{c}{\textbf{Max}} & \multicolumn{2}{c}{\textbf{None}} \\
\cline{2-9}
& \multirow{2}{*}{\textbf{Same}} & \textbf{Halved} & \multirow{2}{*}{\textbf{Same}} & \multirow{2}{*}{\textbf{Halved}} & \multirow{2}{*}{\textbf{Same}} & \multirow{2}{*}{\textbf{Halved}} & \textbf{Same} & \multirow{2}{*}{\textbf{Halved}} \\
& & {\bf (Ours) } &  &  &  &  & { \bf (Expert) } &  \\
\midrule
\textbf{Space} & 1283.7 & 1325.4 & 1384.1 & 1181.6 & 763.5 & 952.0 & 1388.2 & 917.5 \\
\textbf{Invaders} & $\pm$ 728.4 & $\pm$ 504.6 & $\pm$ 622.7 & $\pm$ 424.8 & $\pm$ 193.7 & $\pm$ 304.7 & $\pm$ 351.7 & $\pm$ 385.6  \\
\midrule
\multirow{2}{*}{\textbf{Enduro}} & 1778.0 & 1402.5 & 1723.8 & 1644.3 & 1094.8 & 717.5 & 1428.6 & 1706.2 \\
& $\pm$ 515.9 & $\pm$ 342.8 & $\pm$ 573.9 & $\pm$ 429.6 & $\pm$ 203.3 & $\pm$ 180.9 & $\pm$ 325.9 & $\pm$ 490.7 \\
\midrule
\textbf{Beam} & 6873.3 & 5602 & 8936.0 & 8201.1 & 7871.2 & 4284.0 & 8520.1 & 8032.6 \\
\textbf{Rider}& $\pm$ 2899.6 & $\pm$ 3336.1 & $\pm$ 3636.5 & $\pm$ 3730.4 & $\pm$ 2448.7 & $\pm$ 1685.1 & $\pm$ 2968.6 & $\pm$ 2822.9 \\
\bottomrule
\end{tabular}
\label{tab:ablation}
\end{table}

We perform a detailed ablation study examining the effects of knowledge distillation using Actor-Mimic, halving the number of feature maps in the convolutional layers and doing a spatial max pool over the last convolutional layer independently. The results are summarized in Table \ref{tab:ablation}. We obtain each of the results shown in Table \ref{tab:ablation} similarly to the results presented in Table \ref{tab:main-results}. Specifically, the same experts are used as in our results section where applicable. Each of the models (except for the experts) is trained for 20 million iterations, and evaluated every 0.5 million iterations. The best models were then evaluated on 100 episodes, and the mean and standard deviations of the scores so obtained are reported.

In Table \ref{tab:ablation}, "same" indicates that the number of feature maps have not been reduced, while "halved" indicates that the number of feature maps have been halved (with respect to the architecture of \citet{DQN}). "Max" indicates that global max pooling has been applied, and "None" indicates that it hasn't. "Knowledge Distillation" indicates that training was done using Actor-Mimic and a trained expert agent, while "No KD" indicates that the model has been trained by Q-learning without the use of an expert. Thus, the results in the column "No KD - Max - Halved" correspond to our proposed formulation which uses half the number of feature maps as a DQN, applies global max pooling after the final convolutional layer, and trains the network using Actor-Mimic and an expert. Likewise, the column "Knowledge Distillation - None - Same" represents the DQN expert. Both these results are the same as the results for the corresponding games shown in Table \ref{tab:ablation}

We observe that knowledge distillation is critical for training compressed models. For example in the case of Space Invaders we see that without knowledge distillation, the compressed models perform very poorly (763.5, 917.5 and 952.0 for compressed models vs. 1388.2 for full parameter model). 

\FloatBarrier

However with knowledge distillation the compressed models perform on par (1283.7, 1325.4 and 1181.6 for compressed models vs 1388.2 for the full parameter model). 

We observe a very similar trend in the case of Enduro, where the compressed models trained without knowledge distillation are worse than the baseline (1094.8, 1706.2, 717.5 for the compressed models vs 1428.6 for the baseline). Surprisingly the compressed models trained with knowledge distillation outperform the baseline model (1778.0, 1402.5, 1644.3 for compressed vs. 1428.6 for the baseline)

In the case of BeamRider, we observe that the game is inherently hard and is not amenable to a large degree of parameter reduction. Halving the feature maps has little effect on the performance (8032.6, 8201.1 for models with half number of feature maps vs. 8520 for the baseline model). However applying a global max pool drastically reduces the performance, this drop in performance could not be controlled by knowledge distillation.

\section{Related Work}

The success of Deep Q networks \citep{DQN} on the suite of games in the Arcade Learning Environment \citep{ALE} has sparked an active interest in Deep Reinforcement Learning. \cite{van2016deep}, \cite{dueling} and \cite{schaul2015prioritized} showed significant improvements over the original DQN and achieve superhuman performance on a majority of the games. However to easily and clearly analyze the effect of our techniques for the purpose of model compression, we stick to a simple baseline of the basic DQN architecture and training proposed by \citet{DQN}.

\citet{hinton2015distilling} first proposed that it might be easier for smaller networks to mimic the outputs of a bigger network instead of learning the original task from scratch. They proposed this technique of knowledge distillation where the knowledge of the bigger network (expert) was distilled/transferred into the smaller network (student).

The idea of knowledge distillation has been applied to the domain of deep reinforcement learning to train agents which can play multiple games simultaneously. \citet{parisotto2015actor, rusu2015policy} are notable attempts at exploring the above. Both works learn constrained student networks which are forced to mimic the output of the unconstrained expert networks. The output is either the policy directly generated by the expert network, or a policy inferred from the Q values generated by the network. Similar to the work of \citet{ross2011reduction}, we sample actions from the student network, while imitating the expert's behaviour. The above works train the student network to multitask. In contrast, we study the effectiveness of applying the same techniques when the student network is parameter efficient.

Adding interpretability to the predictions of deep neural networks has been an actively studied area of research. A lot of works have focused on localizing the regions in the input which are most relevant while making a prediction as a proxy for enhancing interpretability. These localized regions of importance can help one understand and debug the decision of DNNs in some cases. \citet{selvaraju2017grad, zhou2016learning} are some notable works which induce weak localization of entities most relevant for making the predictions. We on the other hand follow the approach more in line with \citet{oquab2015object}, which demonstrates that using a global max pool operation forces the network to localize the classified entities. In our work we employ a global max pool to simultaneously induce weak localization and reduce the number of parameters.

\section{Conclusion}

In this paper, we explored the potential of using model compression in the context of Deep RL. In particular, we do this by exploring the applicability of Actor-Mimic to train a light-weight student network from an expert agent. We achieve compression both by reducing the number of feature maps in the student, and by applying a global max pool between the final convolutional layer and the fully connected layer; thereby cutting the number of parameters to less than 3\% of the expert network, while achieving comparable performance. We then show an added advantage of the global max pool formulation: the global max pool allows for the agent to learn to localize important aspects in the environment in a weakly supervised manner, allowing for the visualization of these objects. Finally, we perform an ablation study showing the utility of the different parts of our proposed formulation.

There has recently been a lot of work on the compression of, and efficiency of parameters in, deep learning networks; for example, by using pruning (as in \citet{han2015deep,guo2016dynamic,ashok2017n2n}) or new network architectures (as in \citet{iandola2016squeezenet,howard2017mobilenets}). An interesting line of future work might be to explore the applicability of these techniques and architectures in the context of compression in reinforcement learning.

\section*{Acknowledgments}

We would like to thank Emilio Parisotto for his helpful feedback.

\FloatBarrier
\bibliography{main}
\bibliographystyle{achemso}

\end{document}